\documentclass[]{article}

\usepackage{graphicx,float}
\usepackage{subcaption}
\usepackage{amsmath,amsbsy,bm,amssymb}
\usepackage[hidelinks,colorlinks=false,pdfborder={0 0 0}]{hyperref}
\usepackage{cleveref}
\usepackage{array}
\usepackage{algorithm}
\usepackage{algpseudocode}
\usepackage{etoolbox}
\usepackage{tikz}
\usepackage{enumitem}
\usepackage{diagbox}
\usepackage[group-minimum-digits=2,detect-weight=true,detect-family=true,group-separator={,}]{siunitx}
\usepackage[left=2.75cm,right=2.75cm,top=3.25cm,bottom=3.25cm,includeheadfoot]{geometry} 

\crefname{figure}{Figure}{Figures}
\crefname{table}{Table}{Tables}
\crefname{algorithm}{Algorithm}{Algorithms}
\crefname{equation}{Eq.}{Eqs.}
\Crefname{figure}{Figure}{Figures}
\Crefname{table}{Table}{Tables}
\Crefname{algorithm}{Algorithm}{Algorithms}
\Crefname{equation}{Eq.}{Eqs.}

\begin{document}


\newcommand{\articleauthor}[3]{\href{https://orcid.org/#3}{#1\textsuperscript{#2}}}
\title{The Good, the Bad and the Ugly: \\ Augmenting a black-box model with expert knowledge}
\author{\articleauthor{Raoul Heese}{1,2,*}{0000-0001-7479-3339},
	\articleauthor{Micha{\l} Walczak}{1,2}{0000-0001-5151-7462},
	\articleauthor{Lukas Morand}{3}{0000-0002-8566-7642},
	\articleauthor{Dirk Helm}{3}{0000-0002-5398-5443}, 
	\articleauthor{Michael Bortz}{1,2}{0000-0001-8169-2907}}
\date{%
	\textsuperscript{1} Fraunhofer Center for Machine Learning\\%
	\textsuperscript{2} Fraunhofer Institute for Industrial Mathematics ITWM\\%
	\textsuperscript{3} Fraunhofer Institute for Mechanics of Materials IWM\\%
	\textsuperscript{*} \href{mailto:raoul.heese@itwm.fraunhofer.de}{raoul.heese@itwm.fraunhofer.de}
}

\maketitle

\begin{abstract}
We address a non-unique parameter fitting problem in the context of material science. In particular, we propose to resolve ambiguities in parameter space by augmenting a black-box artificial neural network (ANN) model with two different levels of expert knowledge and benchmark them against a pure black-box model.


\end{abstract}

\section{Introduction}
A central aspect of the physical description of elastoplastic solids are stress-strain relationships. They describe the deformation behavior of a material as a reaction to an external load. In this manuscript, we consider the exponential hardening model \cite{helm2006}
\begin{equation} \label{eqn:R}
	R(\varepsilon,\mathbf{p}) = \frac{\gamma_1}{\beta_1} ( 1 - e^{- \beta_1 \varepsilon} ) + \frac{\gamma_2}{\beta_2} ( 1 - e^{- \beta_2 \varepsilon} ).
\end{equation}
The hardening stress $R$ depends on the accumulated plastic strain $\varepsilon$ and the material model parameters $\mathbf{p} \equiv (\gamma_1,\gamma_2,\beta_1,\beta_2)$.\par
The so-called \emph{parameter identification problem} \cite{mahnken2004} entails finding material parameters $\mathbf{p}$ for a given stress-strain curve $\mathbf{C} = \{ R_1( \varepsilon_1),$ $\dots,$ $R_S(\varepsilon_S) \}$ evaluated at a discretized strain interval $\varepsilon_1,\dots,\varepsilon_S$. Thus, to solve this problem, we search for a suitable inverse mapping
\begin{equation} \label{eqn:E}
	E: \boldsymbol{\mathcal{C}} \longmapsto \boldsymbol{\mathcal{P}} \hspace{.5cm} \text{with} \hspace{.5cm} E(\mathbf{C}) = \mathbf{p}
\end{equation}
from the domain of stress-strain curves $\boldsymbol{\mathcal{C}}$, to the domain of material parameters $\boldsymbol{\mathcal{P}}$. A summary of typical solution strategies can be found in Ref.~\cite{mahnken2004} and references therein. However, all of these approaches only consider unambiguous backward mappings. In contrast, \cref{eqn:R} is symmetric with respect to the permutation $(\gamma_1 \mapsto \gamma_2,\gamma_2 \mapsto \gamma_1,\beta_1 \mapsto \beta_2,\beta_2 \mapsto \beta_1)$ and therefore non-injective, which makes \cref{eqn:E} ambiguous.\par
In the following, we will present three different models for \cref{eqn:E} based on ANNs and compare their performance. In particular, we will demonstrate that incorporating expert knowledge about the forward mapping into the model can improve the results.

\section{Data}
The test and training data consists of stress-strain curves $\mathbf{C} \in \boldsymbol{\mathcal{C}}$ generated from the forward mapping $R(\varepsilon,\mathbf{p})$ with the same equally-spaced discretization for $S=20$. We use $\mathbf{C}_\mathbf{p}$ to denote the stress-strain curve for the parameters $\mathbf{p} \in \boldsymbol{\mathcal{P}}$. Since a uniform sampling in $\boldsymbol{\mathcal{P}}$ will lead to a very unbalanced distribution of curves in $\boldsymbol{\mathcal{C}}$, we choose our data in such a way that all curves of the sample have an approximately even distance to their nearest neighbor in $\boldsymbol{\mathcal{C}}$. For this purpose, we define the distance between two curves $\mathbf{C}_\mathbf{p}$ and $\mathbf{C}_\mathbf{p'}$ as
\begin{equation} \label{eqn:d}
	d(\mathbf{C}_\mathbf{p}, \mathbf{C}_\mathbf{p'}) \equiv \frac{1}{\varepsilon_S - \varepsilon_1} \int_{\varepsilon_1}^{\varepsilon_S} \left| R(\varepsilon, \mathbf{p}) - R(\varepsilon, \mathbf{p'}) \right| \mathrm{d} \varepsilon.
\end{equation}
Summarized, our test and training data contains curves sampled for parameters in the sets $\boldsymbol{\mathcal{P}}_{\mathrm{test}} \subset \boldsymbol{\mathcal{P}}$ and $\boldsymbol{\mathcal{P}}_{\mathrm{train}} \subset \boldsymbol{\mathcal{P}}$, respectively.

\section{Models}
\begin{figure}
\begin{subfigure}[]{0.49\textwidth}
                \centering\includegraphics[scale=1]{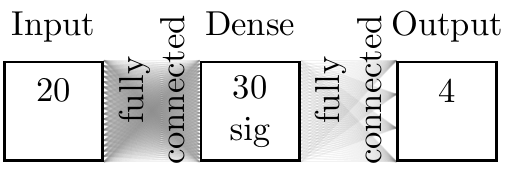}
                \caption{}\label{fig:mlp}
\end{subfigure}%
\begin{subfigure}[]{0.49\textwidth}
                \centering\includegraphics[scale=1]{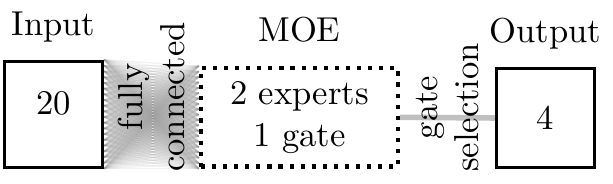}
                \caption{}\label{fig:mdn}
\end{subfigure}%
\caption{ANN architectures. (a) For the naive black-box model $\smash{\hat{E}_{\mathrm{bad}}}$ and the grey-box model $\smash{\hat{E}_{\mathrm{good}}}$ consisting of three fully-connected layers. In total, there are \num{754} trainable model parameters for each of the two models. (b) For the enhanced black-box model $\smash{\hat{E}_{\mathrm{ugly}}}$ with \num{1988} trainable model parameters. Each expert consists of one dense layer with \num{30} neurons and a $\tanh$ activation function, the gate has a single layer with \num{10} neurons and a $\tanh$ activation function.}\label{fig:anns}
\end{figure}

We consider three models with different network architectures, each of them based on a different perspective on the problem. The input of all models is the twenty-dimensional vector of hardening stresses $(R_1, \dots, R_{20})$ associated with a stress-strain curve $\mathbf{C}$. Their output is a four-dimensional parameter vector $\mathbf{p}$.
\begin{enumerate}[label=\textbf{\arabic*})]
\item \textbf{The bad:} The first model, $\smash{\hat{E}_{\mathrm{bad}}}$, is a fully-connected MLP as shown in \cref{fig:mlp} with a conventional least squares loss function. Such kind of models have already been applied successfully to parameter identification problems with unambiguous backward mappings \cite{mahnken2004}. However, we expect them to fail for our ambiguous problem since the model will in fact be trained to predict the mean of the ambiguities in the data as proven in Ref.~\cite{bishop1994}. We therefore regard this model as a naive black-box approach \cite{kroll2000} to the problem, which can be used as a worst-case limit for the other models. Thus this model serves as the ``bad'' candidate among our competitors.
\item \textbf{The good:} The second model, $\smash{\hat{E}_{\mathrm{good}}}$, is a fully-connected MLP of the same architecture as $\smash{\hat{E}_{\mathrm{bad}}}$, but with a custom loss function
\begin{equation} \label{eqn:loss-custom}
	L_{\mathrm{good}} = \sum_{\mathbf{p} \in \boldsymbol{\mathcal{P}}_{\mathrm{batch}}} \sum_{i=1}^{S} \left[ R(\varepsilon_i, \mathbf{p}) - R(\varepsilon_i, |\hat{E}_{\mathrm{good}}(\mathbf{C}_{\mathbf{p}})|) \right]^2
\end{equation}
depending on the current training batch $\boldsymbol{\mathcal{P}}_{\mathrm{batch}} \subset \boldsymbol{\mathcal{P}}_{\mathrm{train}}$. Thus, our expert knowledge about the forward mapping is directly incorporated into the loss function and allows us to circumvent parameter ambiguities. This model can consequently be seen as a grey-box approach \cite{kroll2000} to the problem. Since MLPs are comparably easy to train and the choice of loss function allows us to exploit expert knowledge efficiently, we consider this model as the ``good'' candidate.
\item \textbf{The ugly:} The third model, $\smash{\hat{E}_{\mathrm{ugly}}}$, is a mixture of experts (MOE) network \cite{yuksel2012} as shown in \cref{fig:mdn} with two experts and one gate. For this model we use expert knowledge about the symmetry of the forward mapping to set the number of experts, but make no further use of the specific form of \cref{eqn:R}. We can therefore consider this model as an enhanced black-box approach. On the one hand, MOE networks can handle complicated data ambiguities and are even able to resolve them in a useful manner. On the other hand, they are much more difficult to train than regular MLP networks due to higher complexity. Therefore, this model represents the ``ugly'' candidate in our benchmark. A similar approach has also been studied in Ref.~\cite{morand2019}.
\end{enumerate}

\section{Results}
To test the model performance, we train each of the three models $N=\num{100}$ times with different random seeds for the network initialization. For each training instance $i$, we calculate the mean prediction error
\begin{equation}
	\delta(E,i) \equiv \frac{1}{|\boldsymbol{\mathcal{P}}_{\mathrm{test}}|} \sum_{\mathbf{p} \in {\boldsymbol{\mathcal{P}}}_{\mathrm{test}}} d(\mathbf{C}_\mathbf{p}, \mathbf{C}_\mathbf{E(\mathbf{C}_{\mathbf{p}})})
\end{equation}
and the maximum prediction error
\begin{equation}
	\Delta(E,i) \equiv \max_{\mathbf{p} \in \boldsymbol{\mathcal{P}}_{\mathrm{test}}} d(\mathbf{C}_\mathbf{p}, \mathbf{C}_\mathbf{E(\mathbf{C}_{\mathbf{p}})})
\end{equation}
for the model $E$ with respect to the test data $\boldsymbol{\mathcal{P}}_{\mathrm{test}}$. We summarize the benchmark results in \cref{tab:bm}, where we show the averages $\langle \delta(E,i) \rangle$ and $\langle \Delta(E,i) \rangle$ and the standard deviations $\sigma [\delta(E,i)]$ and $\sigma [\Delta(E,i)]$, respectively, over all $N$ training instances for all three models. We also list the best instance results $\min_i \delta(E,i)$ and $\min_i \Delta(E,i)$. Our implementation is realized with the help of Ref.~\cite{tensorflow}.

\begin{table}
\newcolumntype{S}{>{\centering\let\newline\\\arraybackslash\hspace{0pt}}m{1.45cm}}
\newcolumntype{B}{>{\centering\let\newline\\\arraybackslash\hspace{0pt}}m{1.75cm}}
\centering
\caption{Benchmark results: performance of the three models in solving the parameter identification problem. Smaller values are better, the best results are highlighted in bold. The grey-box approach $\smash{\hat{E}_{\mathrm{good}}}$ is clearly superior.}\label{tab:bm}
\begin{tabular}{|c|S|S|S|S|B|B|} \hline
\diagbox{Model}{Metric} & $\langle \delta(E,i) \rangle$ & $\langle \Delta(E,i) \rangle$ & $\sigma [\delta(E,i)]$ & $\sigma [\Delta(E,i)]$ ${}$ & $\underset{i}{\min} \delta(E,i)$ & $\underset{i}{\min} \Delta(E,i)$ \\ \hline
\rule{0pt}{10pt} $\hat{E}_{\mathrm{bad}}$ & \num{20.27} & \num{320.77} & \num{2.24} & \num{97.38} & \num{15.57} & \num{155.17} \\ \hline
\rule{0pt}{10pt} $\hat{E}_{\mathrm{good}}$ & \textbf{\num{0.88}} & \textbf{\num{9.48}} & \textbf{\num{0.11}} & \textbf{\num{17.52}} & \textbf{\num{0.73}} & \textbf{\num{6.21}} \\ \hline
\rule{0pt}{10pt} $\hat{E}_{\mathrm{ugly}}$ & \num{15.75} & \num{2397.69} & \num{3.75} & \num{1589.80} & \num{9.49} & \num{663.74} \\ \hline
\end{tabular}
\end{table}

The best instance results describe the performance of the best models for application purposes. With regard to the best mean prediction error $\min_i \delta(E,i)$, the grey-box model $\smash{\hat{E}_{\mathrm{good}}}$ is clearly the best candidate, followed by the smart black-box model $\smash{\hat{E}_{\mathrm{ugly}}}$. Without surprise, the naive black-box model $\smash{\hat{E}_{\mathrm{bad}}}$ comes last. For the best maximum prediction error $\min_i \Delta(E,i)$, $\smash{\hat{E}_{\mathrm{good}}}$ is also superior by a wide margin. However, $\smash{\hat{E}_{\mathrm{ugly}}}$ has a much worse performance than $\smash{\hat{E}_{\mathrm{bad}}}$. Hence, we can assume that $\smash{\hat{E}_{\mathrm{ugly}}}$ is much more difficult to train than $\smash{\hat{E}_{\mathrm{bad}}}$ and particularly prone to overfitting.\par
The other statistics also allow to quantify the difficulty of the training process. Although the best expected mean prediction error $\langle \delta(E,i) \rangle$ of $\smash{\hat{E}_{\mathrm{ugly}}}$ is better than for $\smash{\hat{E}_{\mathrm{bad}}}$, its best maximum prediction error $\langle \Delta(E,i) \rangle$ is much worse, what again indicates overfitting. Additionally, the corresponding variances are much higher for $\smash{\hat{E}_{\mathrm{ugly}}}$ in comparison with the other candidates, which underpins our assumption of a much more difficult training process.\par
Summarized, we find that it can be a very useful strategy to directly incorporate the expert knowledge about the forward mapping of a parameter identification problem in the loss function of the model. This approach simplifies the training process in comparison with a MOE model and furthermore leads to a superior overall prediction performance.\par
The conceptional idea presented here can also be used as an origin for further studies. One could examine in how far the choice of the training and test data sets influences the results. A further point to consider is the robustness to noise in the data (which occurs for experimentally measured stress-strain curves). Finally, the method can also be easily adapted to parameter identification problems in other fields of application.


\bibliographystyle{custom}

\bibliography{literature}

\end{document}